 \documentclass[letterpaper, 10 pt, conference]{ieeeconf}  

\IEEEoverridecommandlockouts                              

\usepackage{graphics} 
\usepackage{epsfig} 
\usepackage{mathptmx} 
\usepackage{times} 
\usepackage{amsmath} 
\usepackage{amssymb}  
\usepackage{bm}
\usepackage{algorithm}
\usepackage{multirow}
\usepackage{verbatim}
\usepackage{graphicx}
\usepackage{subfigure}
\usepackage{color}
\usepackage{url}

\title{\LARGE \bf
A Quantitative Analysis of Activities of Daily Living: Insights into Improving Functional Independence with Assistive Robotics
}

\author{Laura Petrich$^{1}$, Jun Jin$^{1}$, Masood Dehghan$^{1}$ and Martin Jagersand$^{1}$ 
\thanks{$^{1}$L. Petrich, J. Jin, M. Dehghan and M. Jagersand are with Department of Computing Science, 
        University of Alberta, Canada,
        {\tt\small jag@cs.ualberta.ca}}%
}

\begin{document}

\maketitle
\thispagestyle{empty}
\pagestyle{empty}


\begin{abstract}
Human assistive robotics have the potential to help the elderly and individuals living with disabilities with their Activities of Daily Living (ADL). Robotics researchers focus on assistive tasks from the perspective of various control schemes and motion types. Health research on the other hand focuses on clinical assessment and rehabilitation, arguably leaving important differences between the two domains. In particular, little is known quantitatively on which ADLs are typically carried out in a persons everyday environment - at home, work, etc. Understanding what activities are frequently carried out during the day can help guide the development and prioritization of robotic technology for in-home assistive robotic deployment. This study targets several lifelogging databases, where we compute (i) ADL task frequency from long-term low sampling frequency video and Internet of Things (IoT) sensor data, and (ii) short term arm and hand movement data from 30 fps video data of domestic tasks. Robotics and health care communities have differing terms and taxonomies for representing tasks and motions. In this work, we derive and discuss a robotics-relevant taxonomy from quantitative ADL task and motion data in attempt to ameliorate taxonomic differences between the two communities. 
Our quantitative results provide direction for the development of better assistive robots to support the true demands of the healthcare community. 
\end{abstract}

\section{Introduction}
        \label{sec:introduction}
\begin{figure}
    \begin{center}
        \includegraphics[width=0.35\textwidth]{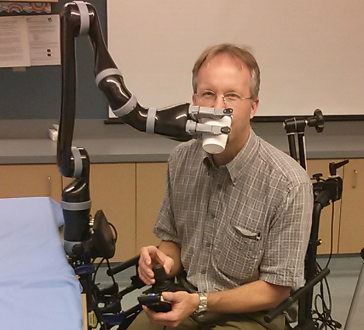} 
        \setlength{\belowcaptionskip}{-8pt} 
        \caption{Jaco assistive arm \cite{Archambault201161Arm}. Wheelchair-mounted assistive robot arms can help those living with disabilities carry out their Activities of Daily Living (ADLs), such as picking up objects, eating, drinking, opening doors, operating appliances, etc.}
        \label{fig:Jaco} 
    \end{center}
\end{figure}

Activities of Daily Living (ADL) can be a challenge for individuals living with upper-body disabilities and assistive robotic arms have the potential to help increase functional independence \cite{Smarr201026Environment}. Assistive robot arms, such as the wheelchair-mountable Kinova Jaco \cite{Archambault201161Arm} (Fig.~\ref{fig:Jaco}) and  Manus/iArm \cite{DriessenMANUSRobot}, have been commercially available for over a decade. Such devices can increase independence, decrease the caregiver load, and reduce healthcare costs \cite{Stuyt200564A}. Robot arms have the potential to be as important to individuals living with upper body disabilities as power wheelchairs have become to those with lower body disabilities. However, outside of research purposes, only a few hundred assistive arms, primarily in Europe and North America, are practically deployed and in use. The gap between assistive robotic research and healthcare needs impedes the wide adoption of assistive robot products. Healthcare professionals, assistive technology users, and researchers have differing biases towards what tasks are of high priority to focus efforts on. For assistive robotics research, knowing which ADLs are most important to support, as well as the necessary performance parameters for these tasks will be crucial to increase usability and deployment. In order to build an assistive robotic task taxonomy that focuses on functional independence, it is imperative to understand how the health care community defines independence; to this end we briefly review the World Health Organization Disability Assessment Schedule (WHODAS2.0) \cite{Ustun2010}.
\cite{WorldHealthOrganizationWHO200339Health}. This classification was primarily developed to determine an individuals level of disability and design an appropriate rehabilitation plan, not to guide assistive robotics research. To the best of the authors knowledge, in healthcare literature there does not appear to be quantitative studies or statistics breaking down individual ADL tasks and motions by able-bodied or individuals living with disabilities.
Health care and robotic domains use different taxonomies to classify and everyday activity tasks and motions \cite{Katz197934Activities,Lawton196936Living,Langdon201604Taxonomy,Bloomfield200332Tasks,Dollar2014ClassifyingLiving,Bullock201105Behavior}. By merging these taxonomies and quantifying health care needs with robotic capabilities we seek to bridge the two, often separate, communities. This would provide the robotics community with guidance as to which tasks could make a large impact to patient populations if implemented on assistive robotic systems.

\begin{figure}
    \begin{center}
        \includegraphics[width=0.35\textwidth]{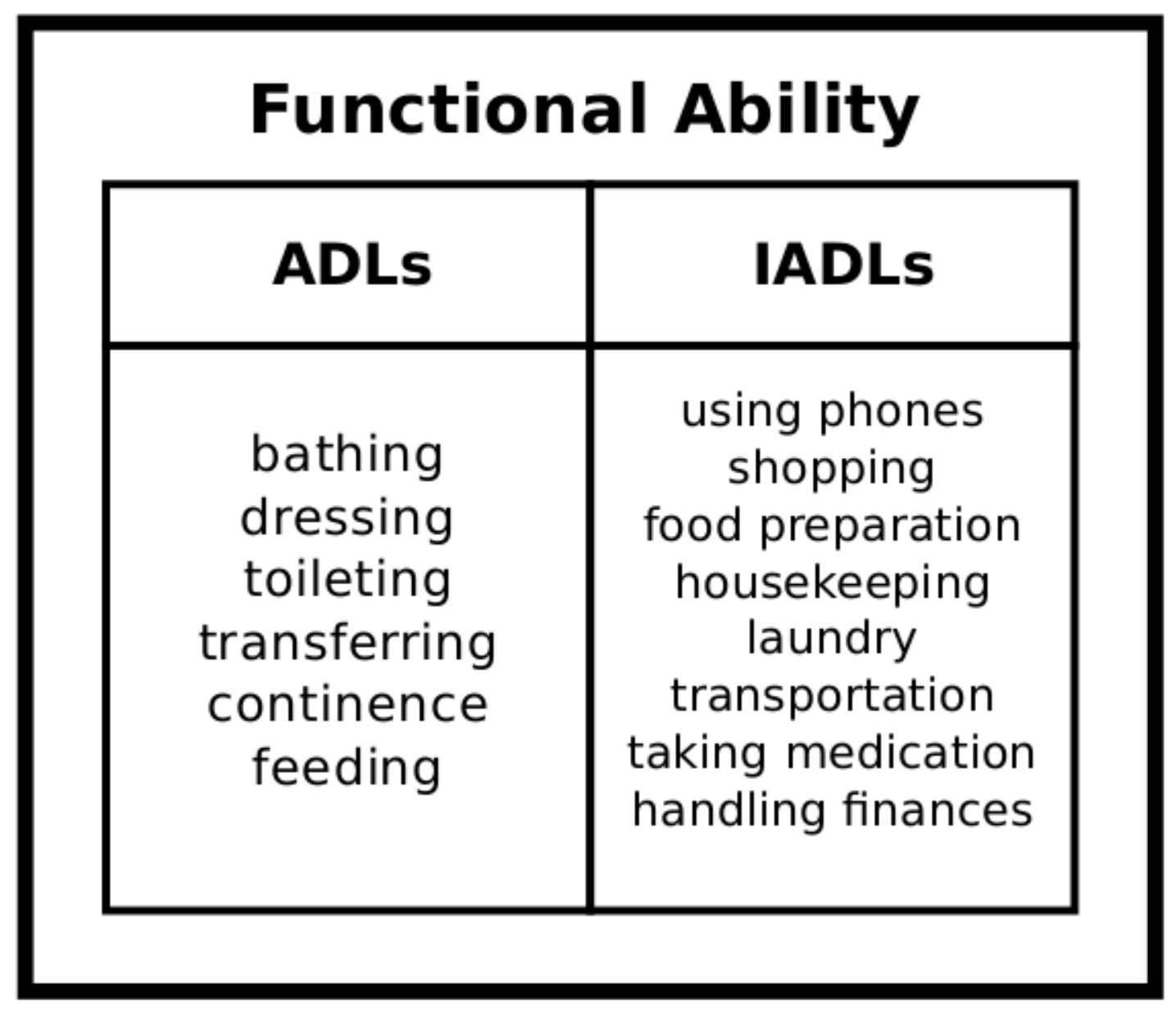} 
        \caption{Measures of Functional Ability that are defined as essential in healthcare communities: Activities of Daily Living (ADLs) \cite{Katz197934Activities} and Instrumental Activities of Daily Living (IADLs) \cite{Lawton196936Living}. ADLs are basic self-care tasks, whereas IADLs are more complex skills. Together they represent what a person needs to be able to manage on their own in order to live independently. }
        \label{fig:FunctionalAbility} 
    \end{center}
\end{figure}

In the field of Computer Science, recent interest in video object and activity recognition \cite{Wang201319Activities, Wang2016} along with life-logging capture has resulted in numerous public data-sets \cite{GurrinCathalJohoHideoHopfgartnerFrankZhouLitingAlbatal201623Research}. In this work we analyzed over 30 such data-sets in order to extract everyday tasks of high importance and relevant motion data \cite{Bolanos201725Overview}.

This paper aims to mitigate the gap dividing the health care and robotics communities. Contributions include:
\begin{enumerate}
    \item We build a task taxonomy consolidating the taxonomic differences between the robotics and healthcare communities for the purpose of further analyzing ADL tasks and the motions they are composed of.
    \item We analyze long term video-recordings from publicly available life-logging data. From the video data we extract ADL task frequencies, that quantify how often a human performs particular ADLs.
    \item From higher frame-rate video recordings of human kitchen activities, we analyze human arm and hand motion data to quantify the speed and variability of human movement. 
    \item We discuss how the task frequency and human motion characterization can prioritize what robotics techniques will have high impact in assistance robotics for elderly and disabled.
\end{enumerate}

\section{Societal and Economic Impacts}
      \label{sec:background}
The use of robotics to help increase functional independence in individuals living with upper limb disabilities has been studied since the 1960’s. We distinguish here between a physically independent robot arm, typically mounted on the users wheelchair, and a smart prosthesis, attached to an amputated limb, with the former being our group of interest. 
The United States Veterans Affairs estimate that approximately 150,000 Americans could benefit from currently commercially available wheelchair-mounted robot arms \cite{Chung201316Review}. With improved functionality, reliability, and ease of use deployment to larger populations could be possible. 



What is the magnitude of need and potential for robotic assistance in the world? Many countries in the west and Asia have an aging populations and disabilities also affect younger populations, e.g. from accidents, disease, or inheritance. Definitions and quality of statistics on disability differs across nations and are difficult to integrate globally. Canada has a multi-ethnic population and characteristics similar to other industrialized nations. The proportion of seniors (age 65+) in the population is steadily increasing, with seniors comprising a projected 23.1\% of the population by 2031 \cite{StatCan2}. In 2014, seniors constituted only 14\% of the population, but consumed 46\% of provincial public health care dollars \cite{StatCan3}. A growing number of elderly and disabled, supported by a dwindling young population is putting pressure both on government budgets and available health care personnel. Today, individuals with lower-body impairments and the elderly are able to independently move around using power wheelchairs. In the near future wheelchair-mounted robot arms could help increase independence and reduce care needs for those living with reduced upper-limb function.

Statistics Canada found that from 2001 - 2006 there was a 20.5\% increase in those identifying as having a disability, corresponding to over 2.4 million people in Canada \cite{StatCan}.  
One in twenty Canadians living with disabilities regularly receive assistance with at least one ADL on a daily basis, although not all of which will require the use of wheelchair-mounted arms. This suggests that there is a significant need and potential market for robotic solutions in Canada and similar countries across the world. Some individuals may prefer automation integration with their smart homes, and some may require both cognitive and physical assistance. While artificial intelligence might provide some basic cognitive support, such as planning of the days tasks and reminders, it cannot eliminate the need for human contact and support. However, robotic assistance can free up humans from mundane chores, allowing more time for caregivers to focus on high quality help and personal interaction. A four year study of assistive arm users in care homes found that a robot reduced the nursing assistance need by 40\% from 3.7h/day to 2.8h/day \cite{Romer2005}. 
While cost savings from reduced nursing are already significant (about \$20,000/year in an industrialised economy), further savings and increased independence came from half of the robot users being able to move out of assistive living with one quarter obtaining jobs. 

An advantage of wheelchair-mounted arms is that they are with the person at all times. Nursing care is typically only for morning and evening routines for those who live independently. Imagine yourself  dropping something important and having to wait all day before someone is able to help you retrieve it.

\section{Activities of Daily Living, Self-Care, and Functional Independence}
        \label{sec:adl}
\begin{figure}[t]
    \begin{center}
        \hbox{\hspace{-3em} \includegraphics[scale=0.49]{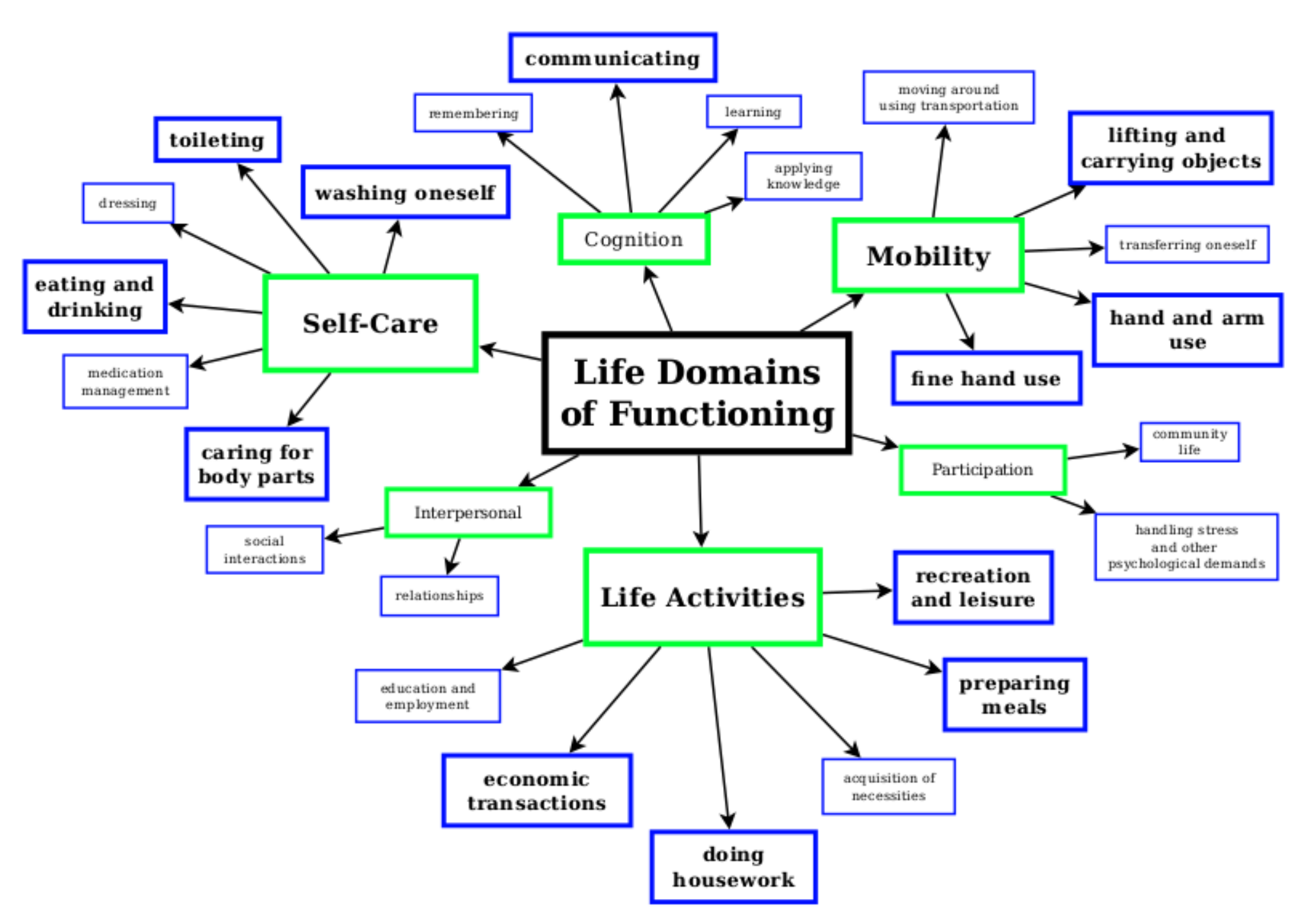}}
        \caption{The major life domains of functioning and disability as set out in the World Health Organization Disability Schedule 2.0 (WHODAS2.0); a standardized cross-cultural measurement of health status based on the International Classification of Functioning, Disability and Health. WHODAS2.0 can be used to measure the impact of health conditions, monitor intervention effectiveness and estimate the burden of physical and mental disorders across all major life domains. Physical motion activities relevant to this paper are highlighted in bold}
        \label{fig:LifeDomains} 
    \end{center}
\end{figure}

The International Classification of Functioning, Disability and Health (ICF) provides a framework for determining the overall health of individuals and populations \cite{WorldHealthOrganizationWHO200339Health}. Disability information is an important indicator of a population’s health status, as it shows the impact that functional limitations have on independence. This concept is known as functional disability, or the limitations one may experience in performing independent living tasks \cite{Spector199837Disability}. A quantification of functional disability includes measures of both Activities of Daily Living (ADLs) \cite{Katz197934Activities} and Instrumental Activities of Daily Living (IADLs) \cite{Lawton196936Living} (Fig. \ref{fig:FunctionalAbility}); in this work we will refer to these collectively as ADLs. The World Health Organization further developed the World Health Organization Disability Assessment Schedule (WHODAS2.0) from the ICF as a standardized, cross-cultural measure of functioning and disability across all life domains \cite{Federici201709Review}. Figure \ref{fig:LifeDomains} highlights these major life domains with associated tasks; the tasks most relevant to robotics research are emphasized in italics. 

A common approach that drives research is to ask patients and caregivers for their preferences when it comes to robotic assistance \cite{Beer201213Place,Stanger199427Priorities}. Notably, preferences vary and user opinions shift over time. In particular, a survey of 67 users surveyed both before and after they received and used an assistive robotic arm found that caregivers tend to favor essential tasks, such as taking medication. Pre-automation patients favor picking up dropped objects and leisure-related tasks, with a shift more towards work-related tasks post-automation \cite{Chung201316Review}. Combining user preferences with quantitative ADL data will be important for robotics researchers to consider when deciding what tasks should be focused on.

\section{A Task Taxonomy for Arm Manipulation}
        \label{sec:taxonomy}
Robotic capabilities can be built bottom-up by designing control methods for individual motions (i.e. motor primitives) which can then be combined to solve specific tasks \cite{Kober2009}. The same motions can potentially be used to solve different ADLs that fall within healthcare taxonomies. Dexterous in-hand manipulation requires different contact configurations and manipulation taxonomies have been developed to compensate for these various configurations  \cite{Bullock201305bManipulationb}. Robot arm manipulation is generally thought of as a 6-DOF Euclidean (end-effector) transform, thus requiring no taxonomy. Contrarily, ADL tasks naturally contain a variety of movements with different DOFs, as well as contact and precision motions. This suggests that an ADL-based taxonomy could help guide the development of control subroutines tailored to those specific requirements and that the composition of such subroutines will be capable of solving a broad variety of tasks.

\begin{figure}[t]
        \hspace{3mm}
        \includegraphics[width=0.45\textwidth]{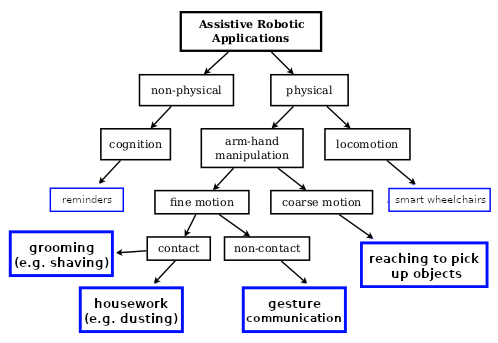} 
        \caption{High-Level Taxonomy of Assistive Robot Tasks and Motions, and how they intersect with example WHODAS tasks from Figure \ref{fig:LifeDomains}.}
        \label{fig:RoboticsTax} 
\end{figure}

Figure \ref{fig:RoboticsTax} introduces a high-level taxonomy of robotic tasks. There are three general categories relevant to assistive robotic applications: non-physical cognitive, locomotion-based mobility, and arm-hand manipulation tasks. In this work we will focus on arm and hand manipulations. In applied robotics, the robot gripper is typically used for grasping while the arm is responsible for gross pose alignment and contact point decisions. There is much work to be done before robotic systems will be able to utilize fine dexterous finger manipulation motions humans use for ADLs. Commonly the robot gripper just grasps and the robot arm has to perform both coarse and fine manipulations \cite{Bullock201105Behavior}.
Coarse reaching motions are mostly a 3-DOF translation and requires moderate accuracy. Fine motions can be further subdivided into contact and non-contact motions depending on the desired outcome. Non-contact 6-DOF fine motions can be used to bring an object into alignment with the target location before putting the object down or inserting it. Although most applied robotics is performed using position-based control, some studies take contact forces into account, either through impedance control or sensing and modeling of the surface for hybrid position-force control \cite{HybridVisionForce}. Surface contact data allows for human-like control strategies to overcome sensing and actuation inaccuracies by utilizing practices such as feeling a table and sliding across it before picking up a small object.

\section{Wheelchair-mounted assistive arms}
        \label{sec:robotic_review}
A lightweight robotic arm can be attached to a wheelchair to assist with ADLs \cite{Chung201316Review}. With such a device, users with limited upper limb functionality are able to independently carry out a larger subset of their daily tasks. While there are about 2 million robot arms deployed in industry, only two assistive robot manufacturers have over 100 assistive arms deployed with disabled humans, namely, Exact Dynamics (Manus and iARM) \cite{DriessenMANUSRobot} and Kinova (JACO and MICO) \cite{Campeau-lecours2018KinovaApplications}. These arms are lightweight with integrated controllers and cost around USD 20,000-35,000 with a gripper. For example, the Kinova JACO robotic arm weighs 5.7kg (12.5lbs) and comes with a wheelchair attachment kit. It is capable of grasping and moving objects up to 1.5kg, Fig.~\ref{fig:Jaco}. The Manus/iARM has similar specifications. 

In published assistive robotics research a variety of commercial robot arms are used and several new prototype arms have been designed, however neither new robots nor new methods for motion control or Human Robot Interaction (HRI) have reached noticeable deployment \cite{Groothuis201311Arms}. The few hundred deployed JACO and Manus arms still use basic joystick position-based teleoperation, where a 2 or 3 Degree of Freedom (DOF) joystick is mapped to a subset of the Cartesian arm translation and wrist rotation controls \cite{Archambault201161Arm, Goodrich200706bSurvey}. To complete 6-DOF tasks the user needs to switch between various Cartesian planes, known as mode switching, which can be tedious and cognitively taxing.

Novel user interfaces have been implemented in research settings and rely on a variety of input signals for shared autonomy, such as gestures, eye gaze, electromyography (EMG), electroencephalography (EEG), and electrocorticographic (ECoG). Gesture-based systems allow the user to specify an object to manipulate by either pointing \cite{QuinteroRJ15} or clicking on it through a touch screen interface \cite{Jagersand1995,Gridseth2016} and then the robotic arm would autonomously move towards the target object \cite{Tsui201153Arm}. Eye gaze can be used in place of gestures to capture an individuals intended actions and drive robot control \cite{Admoni2016}. Neural interface systems (i.e. ECoG and EEG) work by mapping neuronal activity to multidimensional control signals that are used to drive robot arm movements \cite{Hochberg2013}. Hochberg \textit{et al.} highlight the potential of ECoG-based control methods, although it requires an invasive surgical procedure in order to implant the microelectrode array. EEG- and EMG-based methods provide an intuitive, non-invasive alternative for closed-loop robot control using brain and muscle activity \cite{Gomez2017, Liarokapis2012}. Recently, Struijk \textit{et al.} developed a wireless intraoral tongue interface device that enables individuals with tetraplegia to use their tongue to control a 7-DOF robotic arm and 3 finger gripper \cite{Struijk2017}.

\section{ADL Evaluation from Lifelogging Data}
        \label{sec:lifelogging}
Lifelogging data is a valuable source of quantitative ADL and human motion information. Lifelogging involves long-term recording of all activities performed by an individual throughout the course of the day, usually through a video camera, and occasionally using other types of sensors \cite{GurrinCathalJohoHideoHopfgartnerFrankZhouLitingAlbatal201623Research}. While lifelogging research has been published for over two decades \cite{Mann1997}, hardware and method innovation has made the field grow greatly within the past few years \cite{Bolanos201725Overview}. Small, wearable cameras, such as the Microsoft Lifecam \cite{Wilson2016}, with a longer recording duration has made it more practical compared to the analog video cameras and recorders used in initial research. 
New methods for recognizing objects and actions has driven Computer Vision (CV) research interests to explore lifelogging data, which has been found to be a source of more realistic “in-the-wild” data than typical CV benchmarks \cite{Pirsiavash201214Views, Fathi201122Activities}.

In this work we evaluated over 30 lifelogging datasets\footnote{For a detailed table of specific datasets investigate please visit http://webdocs.cs.ualberta.ca/~vis/ADL/},
most of which targeted the performance of a particular algorithm (e.g. video object recognition in home environments) and therefore did not encompass the full day. These datasets typically did not have a statistically sound sampling over all objects and tasks in order to meet our analysis inclusion criteria for this work. We found that long term video recordings of several days or more were done at 1-2 frames per minute (fpm), making these data useful to analyze gross ADL task frequency and duration, but not suitable for studying detailed timing of individual arm and hand motions. An additional downfall of the low fpm video datasets is that they fail to capture daily tasks which are repeated with high frequency but are performed quickly, such as opening doors or turning on lights. Another category of datasets had regular video rate recordings of specific tasks, at 30 frames per second (fps), making the detailed timings of individual arm and hand motions possible. We were able to choose three sources of data for analysis: two from long duration recordings in order to extract ADL task frequency and duration \cite{GurrinCathalJohoHideoHopfgartnerFrankZhouLitingAlbatal201623Research,Tapia200446Sensors}, and 
one from short-term recordings of individual tasks \cite{Fathi201221Gaze}.

\begin{figure}
    \begin{center}
        \includegraphics[width=0.5\textwidth]{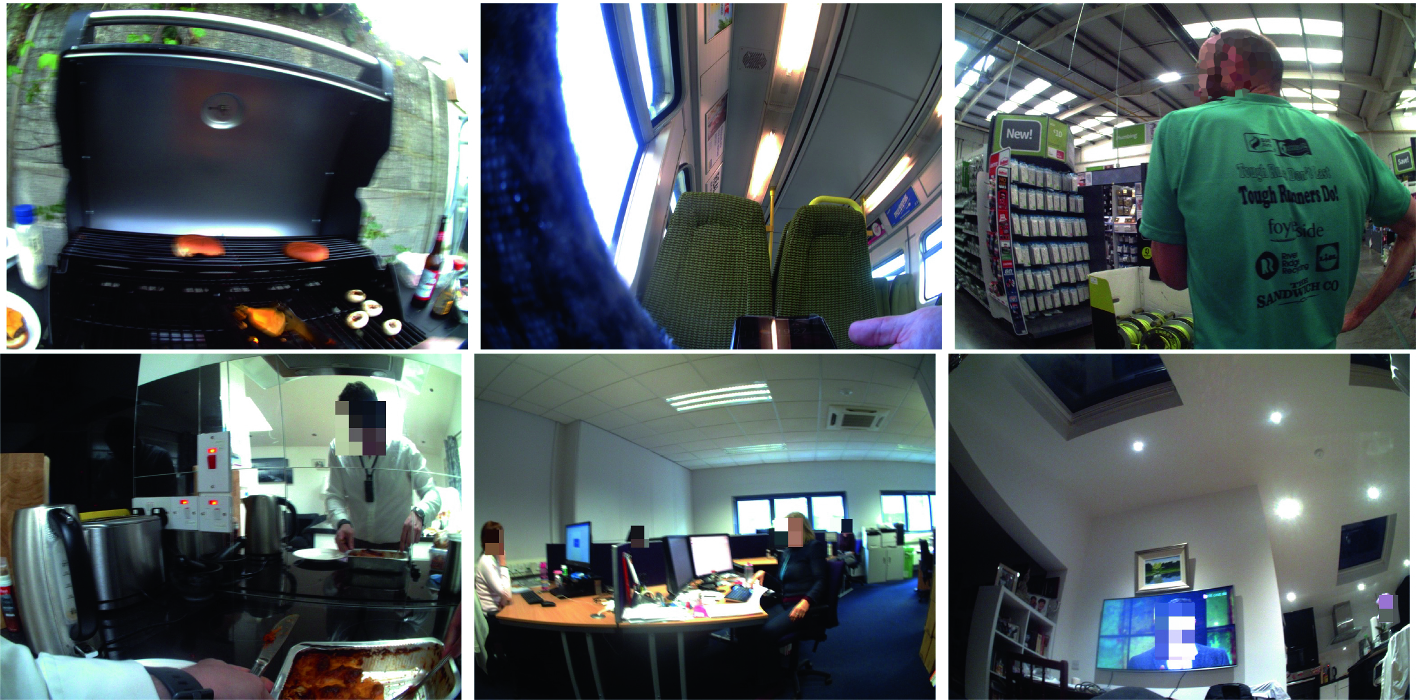} 
        \caption{In the NCTIR Lifelog Dataset \cite{Gurrin201724Task} 3 people wore lifelogging cameras for a total of 79 days, collectively. These provide images of the individuals arms and egocentric environment at a rate of 2 fpm. Due to the low frame rate, fine arm and hand motions are not available, but actions are instead inferred from context using visual concepts automatically computed from the images.}
        \label{fig:NCTIRimages} 
    \end{center}
\end{figure}

\subsection{ADL Task Frequency Analysis}
\begin{figure}
    \begin{center}
        \includegraphics[width=0.5\textwidth]{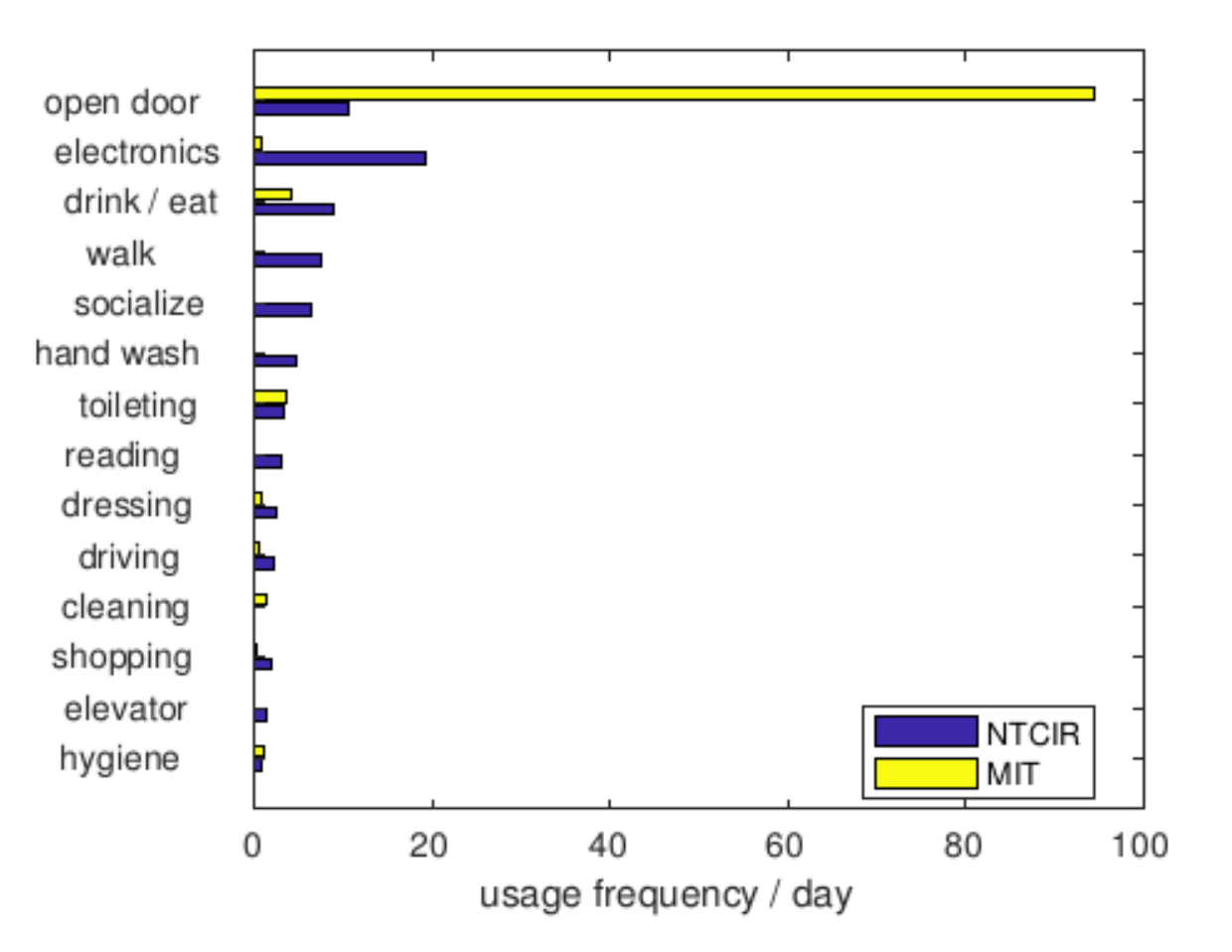} 
        \caption{Human ADL task frequencies from MIT IoT sensors (yellow bars), and NTCIR lifelogging video (blue bars). The largest bar measure is the most accurate, as explained in the text. The 2 frame/minute video analysed from NCTIR misses fast cabinet door and drawer openings to retrieve objects, so under counts doors. MIT under counts electronics, as mobile devices were not sensored. Door and drawer opening and robot feeding are high priority tasks robot researchers already publish on, while hand washing is high priority task where robot assistance has not been studied.}
        \label{fig:ADLFreq} 
    \end{center}
\end{figure}

To compute quantitative data on ADL task frequency and duration we analyzed both egocentric lifelogging videos (referred to as ‘NTCIR’ \cite{GurrinCathalJohoHideoHopfgartnerFrankZhouLitingAlbatal201623Research, Gurrin201724Task}), and exocentric data from Internet-of-Things type sensing built into home objects (referred to as ‘MIT’) \cite{Tapia200446Sensors}. Example lifelogging images from the NTCIR dataset are shown in Fig. \ref{fig:NCTIRimages}. The use of complementary sensing turned out to be important for capturing a broader set of tasks. Similar to other CV research, we were able to infer actions from automatically computed visual concepts \cite{Fathi201221Gaze}. Our supplementary web page (footnote 1) contains the visual context to actions inference bindings, so readers can replicate results or add other rules and actions to classify. We hand-labeled a small portion of the data to verify the accuracy of the automatic computations. This enabled us to label in-home data sequences spanning multiple days according to what ADLs were carried out at particular times and compute their statistics. Figure \ref{fig:ADLFreq} illustrates the frequency of the most common ADL tasks found in these datasets.

We have grouped tasks together that correspond with robot skills rather than specific healthcare ADL/ICF codes. Some events are detected more reliably by the embedded sensors used in MIT, others only in the lifelogging videos. For examples sensors detect quick events more reliably that the lifelogging video data misses. In contrast, outdoor activities are only captured in the video data. By combining results from both datasets, we were able to obtain a better quantitative measure of task significance.

Opening and closing doors is the most frequent task at 94 times per day; this category includes doors between rooms, cabinet doors and drawers. Our rationale for including cabinet doors and drawers is that the robot would approach each situation in the same fashion as a standard door. We believe the MIT data was more accurate since the ‘door opening’ data was obtained from built in door sensors; the low video frequency (2 fpm) of the NTCIR data presented low accuracy with the automatic visual concepts extraction by missing quick openings, particularly of cabinet doors and drawers to retrieve objects. Following door opening, electronics is the second most frequent task performed during the day; the electronics category refers to the use of electronic handheld devices and was dominated by smart phone use. These devices were mostly not covered by the MIT sensors, but were detected in the NTCIR video data. Drinking and eating were essential tasks in both studies, with a frequency of 8.8/day from NTCIR and 4.4/day from MIT. MIT-data captured hand washing by the number of faucet openings/closing (ie. turning the sink on and off resulted in two tasks), which overestimated hand washing frequency. We removed this outlier and relied on the NTCIR results of 4.7/day. These results capture the actions of able adults, and hence can guide robotics researchers both what to implement, and how - a task that is frequent and executed quickly by a human such as door openings need to be easy and fast for a disabled to do with their robot. This depends on the physical velocity of the robot, as well as the time and cognitive load it takes the user to handle the human-robot interface. Door openings are covered in the literature e.g. \cite{}, 
and robot feeding has been studied for over 30 years, with some prominent recent results \cite{Beer201213Place,Gordon2019}. By contrast, hand washing would also be high-priority. Hand washing has been studied in assistive Computer Vision \cite{Hoey2012}, to prompt Alzheimer patients though the steps, 
but we know of no robotics researchers to have attempted this highly important ADL. Yet, we know anecdotally that disabled users of robot arms use the robot to support their own arm (please see the accompanying video). Hence, it should be possible to study motion programming where a robot arm brings the human arm and hand under a water tap (the water tap can be automatically activated as is already common). 


\subsection{Arm and Hand Motion Analysis.}
\begin{figure}
    \begin{center}
        \includegraphics[width=0.5\textwidth]{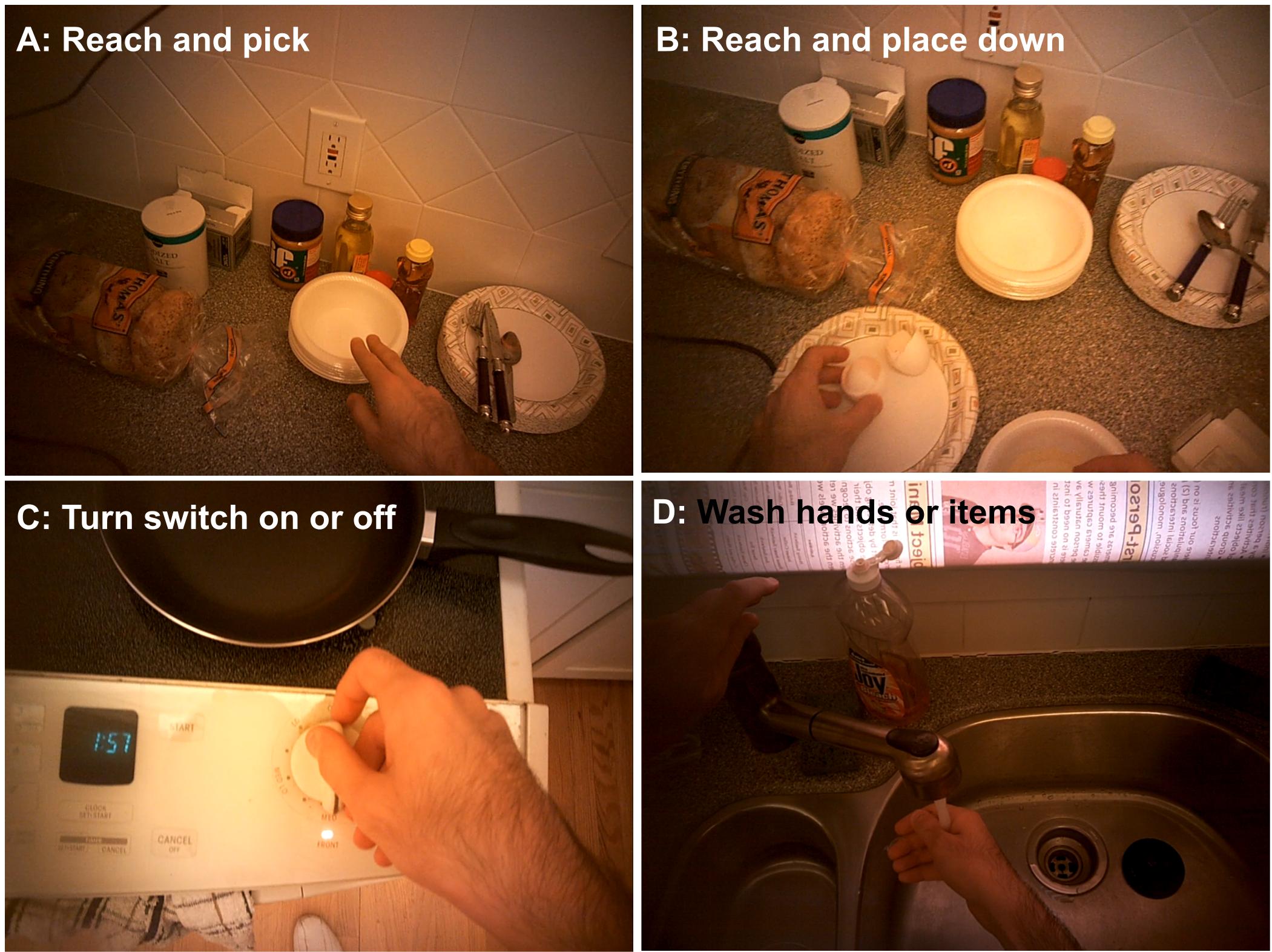} 
        \caption{The GTEA Gaze+ dataset contains 7 kitchen activities performed by 6 persons. We analyzed the frequency and mean human execution time of each human activity. Top 4 frequent activities are as shown above.}
        \label{fig:GTEAimage}
    \end{center}
\end{figure}

From high frame-rate video datasets we were able to extract the number and timings of individual arm and hand motions required to perform a particular ADL and, for a few tasks, similar timings for robot execution. The Georgia Tech Egocentric Activity Datasets (GTEA Gaze+) \footnote{http://www.cbi.gatech.edu/fpv/} contain full frame rate (30 fps) video recordings of humans performing domestic tasks \cite{Fathi201221Gaze}. We analyzed the annotated GTEA Gaze+ dataset, which contained approximately 25GB of annotated kitchen activity videos to extract individual human motion timings performed during these tasks (Fig. \ref{fig:GTEAimage}).

Figure \ref{fig:GTEAmotions} illustrates four common motions out of the 33 captured in the GTEA Gaze+ dataset. Notably, human motions were far faster than typical assistive robot motions. For example, as seen in the histogram, reach motions that take just a second for a human, can take anywhere from ten seconds to several minutes in published HRI solutions  \cite{Muelling201547Manipulation}. This has implications for how many tasks a robot system can practically substitute in a day without taking up an excessive amount of time. In other motions, such as pouring liquids, the task itself constrains the human to proceed rather slowly. The door task covers both lightweight cabinet doors and drawers, along with heavier doors (e.g. refrigerator); with lighter doors, the human times approached that of an unconstrained reach, despite the more challenging physical constraint of hinged or sliding motion, while heavier doors represent the long tail of the time distribution. Unlike NTCIR, GTEA Gaze+ is not a representative sampling of all human activities. It is still notable that the number of reaches is three times the number of door openings (1800 reaches versus 600 door and drawer openings over 11 hours of video). 

In the following table the frequency (occurrences per hour) and mean human task execution time are presented. The tasks involve kitchen activities - food preparation, but movement times are likely typical of other human activities. It is notable how quickly human moves and how many movements we make. Replicating human motion speed and agility is a gold standard to benchmark robots against. 
\begin{center}
\begin{tabular}{ |l|r|r| } 
 \hline
Task & freq & time \\
 \hline
  \hline
Reach and pick item & 88 & 1.5s \\
 \hline
Reach and place item & 84 & 1.2s \\
 \hline
Turn switch on or off & 10 & 2.1s \\
 \hline
Wash hands or items & 3 & 6.7s \\
 \hline
Flip food in pan & 2 & 4.9s \\
 \hline
Transfer food (e.g. to plate) & 6 & 8.6s \\
 \hline
\end{tabular}
\end{center}


\section{Discussion}
\label{sec:discussion}
Door opening/closing, drinking/eating, hand washing and toileting would arguably be the most essential to support for assistive robot arm and hand systems, out of all the ADL tasks analyzed in this work. The first three are relatively feasible to accomplish given the payload capacity of current robotic arms. 

Activities such as using electronics (primarily smartphones), socializing, and reading could be physically aided by robot arms, but since these activities are not inherently physical, alternative solutions are possible and can be a simpler and more reliable solution (e.g. hands-free phone use and other computational automation).

Toileting is a high priority task that involves transferring from a wheelchair to the toilet. Assistive arms do not support this, but there are specialized transfer devices - also useful for transfer from beds - that are generally used in health care, and be employed in peoples homes.

Overall, there is great potential for supporting ADLs for those living with disabilities as well as the elderly. Over the past few decades there has been an increasing demand for health care services due to the rising elderly and disability populations \cite{DepartmentofEconomicandSocialAffairs:PopulationDivision2017WorldHighlights}. Assistive robots can help bridge this gap by alleviating the labour burden for health care specialists and caregivers. Furthermore, an assistive robot could help one perform ADL they are otherwise incapable of managing on their own, thus increasing functional independence. 

\begin{figure}
    \begin{center}
        \includegraphics[width=0.5\textwidth]{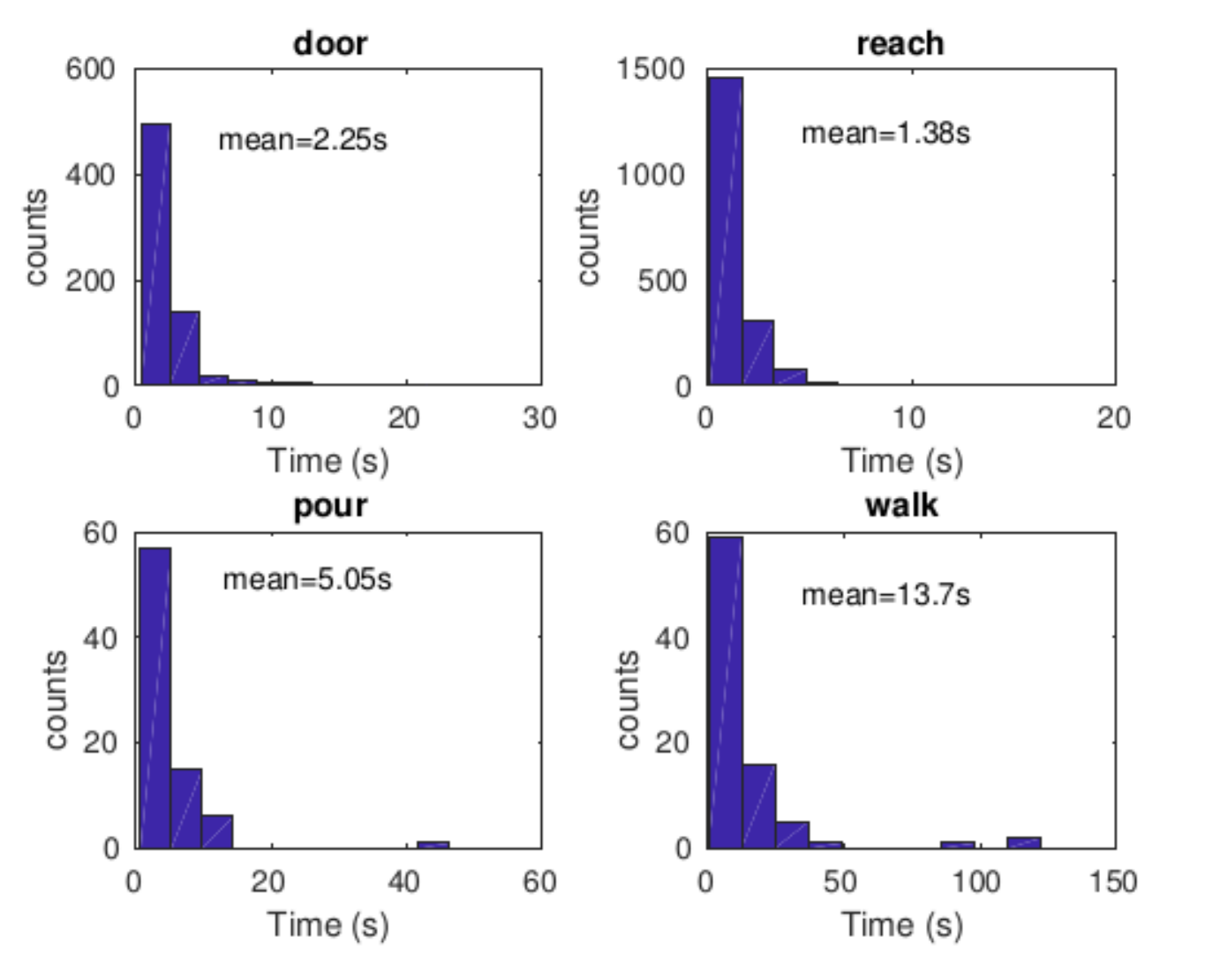} 
        \caption{Timing histograms for four common human motions. Human arm and hand motions are very quick and accurate, just seconds long. By contrast current robots are slow.}
        \label{fig:GTEAmotions} 
    \end{center}
\end{figure}

However, challenges remain before these robots will reach mainstream adoption, including but not limited to: system costs, task completion times, and ease of use via user interfaces. Currently costing around USD 30,000, an arm is a significant expense for an individual, who may already have a limited income. While western health insurance often covers expensive prostheses for amputees, only in the Netherlands does insurance cover a wheelchair mounted arm. 

Speed of robot motion, which affects task completion time, is another challenge. While a human reach takes just 1-2s (Fig.~\ref{fig:GTEAmotions}), published assistive robots take 40-90s, resulting in robot solutions that are magnitudes slower \cite{Kim201252Robot,QuinteroRJ15,Muelling201547Manipulation}. In the GTEA Gaze+ kitchen tasks, humans performed 160 reaches per hour. Substituting robot reaches would turn a moderate 30 minute meal preparation and eating time into a 2 hour ordeal. Anecdotal comments from users of assistive robot arms are that everyday morning kitchen and bathroom activities, which an able person easily performs in less than an hour, takes them several hours.

Robots may solve tasks differently than humans as robots are often limited to grasping one item at a time, while humans can handle many. When setting a table we will for instance pick several utensils at a time from a drawer. In restaurants, waiters can clear a table for four, and handle all the plates, utensils, glasses, etc. in their hands and arms. Analysing the publicly available TUM Kitchen Data Set of activity sequences recorded in a kitchen environment \cite{Tenorth2009}, we found that the robot strategy on average required 1.6 times more movements than a human. Users of assistive robots adopt compromises to deal with the speed and accuracy of robots. For example, foods and drinks that can be consumed while held statically in front of the humans face by the robot, e.g. eating a snack bar, or drinking with a straw, are far quicker to consume than those requiring numerous robot reach motions, such as eating a bowl of cereal.
 
User interfaces need improvements. Currently deployed arms are, as mentioned before, joystick operated, while most research is on autonomous movement, e.g. autonomously delivering a piece of food once the system has detected an open mouth \cite{Park2019, Gordon2019}. Sheridan's conventional scale from tele-operation to autonomy \cite{Sheridan2002}, has been redefined by Goodrich to have seamless human-robot collaboration as the goal rather than robot autonomy \cite{Goodrich200706bSurvey}.

We, and others, have found that users generally prefer to have continuous in-the-loop control \cite{QuinteroRJ15, Kim2012}. Someone may change their mind midway through autonomous food delivery, and may instead open their mouth to say something - only to get their mouth stuffed with food. In very recent work a low dimensional control space is learned from demonstrations. This allows a human user to have direct control over a 6DOF motion using a low DOF HRI, such as a joystick \cite{Jeon2020, Quintero2017}. Getting the balance right between human interaction and semi-autonomous assistive systems will be challenging. Currently, most research is evaluated with a few participants trying it for about an hour each in a research lab setting. We expect that new HRI solutions will need to be deployed longer term in real users homes in order to properly evaluate usability. 

\section{Conclusion}
\label{sec:conclusions}
In this paper we presented assistive robotics for Activities of Daily Living - ADL from both from a health care perspective and robotics perspective. We analyzed human ADL task frequency from public life-logging datasets and computed motion timings from public Computer Vision data. Overall, reach motions (to grasp objects) and door openings (including cabinets and drawers) were the most frequent motions. Drinking, eating and hand washing are other high priority tasks that can be addressed by current assistive robot arms. Toileting and dressing, while ranking just below, are generally thought to be more challenging for robotics, since they require the transfer of body weight. Detailed data on frequency and duration information for all analyzed tasks and motions, as well as the analysis methods are available on the companion website \url{http://webdocs.cs.ualberta.ca/~vis/ADL}


\bibliographystyle{IEEEtran}
\bibliography{References}
\end{document}